%% file: main.tex
\documentclass{article}
\usepackage{ijcai23}

\usepackage{times}
\usepackage{soul}
\usepackage{url}
\def\myurl#1{\setbox0\vbox{\hsize.5\maxdimen
\url{#1}\par
\setbox0\lastbox
\global\setbox1\hbox{\unhbox0\unskip\unskip\unpenalty}}\unhbox1 }

\usepackage[hidelinks]{hyperref}
\usepackage[utf8]{inputenc}
\usepackage[small]{caption}
\usepackage{graphicx}
\usepackage{amsmath}
\usepackage{amsthm}
\usepackage{booktabs}
\usepackage{algorithm}
\usepackage{algorithmic}
\usepackage[switch]{lineno}

\usepackage[many]{tcolorbox}
\usepackage{xcolor,colortbl}
\usepackage{adjustbox}
\usepackage{multicol}
\usepackage{multirow}
\usepackage{tikz}
\usepackage{makecell}
\usepackage{enumitem}

\usepackage{pgfplots}  
\pgfplotsset{width=6.6cm,compat=1.7} 
\usepackage{amssymb}

\title{Evaluating GPT-3 Generated Explanations for Hateful Content Moderation}

\author{
    Anonymous
}

\author{
Han Wang$^{1,*}$
\and
Ming Shan Hee$^{1,*}$\and
Md Rabiul Awal$^2$\and \\
Kenny Tsu Wei Choo$^1$\And
Roy Ka-Wei Lee$^1$
\affiliations
$^1$Singapore University of Technology and Design\\
$^2$Mila - Quebec AI Institute
\emails
han\_wang@sutd.edu.sg,
mingshan\_hee@mymail.sutd.edu.sg,
awalrabiul6@gmail.com, \\
\{kenny\_choo, roy\_lee\}@sutd.edu.sg
}

\begin{document}

\maketitle
\def\thefootnote{*}\footnotetext{Both authors contributed equally to this research.}\def\thefootnote{\arabic{footnote}}


\begin{abstract}
Recent research has focused on using large language models (LLMs) to generate explanations for hate speech through fine-tuning or prompting. Despite the growing interest in this area, these generated explanations' effectiveness and potential limitations remain poorly understood. A key concern is that these explanations, generated by LLMs, may lead to erroneous judgments about the nature of flagged content by both users and content moderators. For instance, an LLM-generated explanation might inaccurately convince a content moderator that a benign piece of content is  hateful. In light of this, we propose an analytical framework for examining hate speech explanations and conducted an extensive survey on evaluating such explanations. Specifically, we prompted GPT-3 to generate explanations for both hateful and non-hateful content, and a survey was conducted with 2,400 unique respondents to evaluate the generated explanations. Our findings reveal that (1) human evaluators rated the GPT-generated explanations as high quality in terms of linguistic fluency, informativeness, persuasiveness, and logical soundness, (2) the persuasive nature of these explanations, however, varied depending on the prompting strategy employed, and (3) this persuasiveness may result in incorrect judgments about the hatefulness of the content. Our study underscores the need for caution in applying LLM-generated explanations for content moderation. Code and results are available at \url{https://github.com/Social-AI-Studio/GPT3-HateEval}.

\end{abstract}

\section{Introduction}
\label{sec:introduction}

\input{introduction}

\section{Related Works}
\label{sec:related}

\input{related}

\section{Methodology and Experimental Setup}
\label{sec:method}
\input{method}

\section{Experimental Results and Analysis}
\label{sec:experiment}
\input{experiments}

\section{Discussion}
\label{sec:discussion}
\input{discussion}

\section{Conclusion}
\label{sec:conclusion}

\input{conclusion}



\bibliographystyle{named}
\bibliography{ref}

\appendix


\end{document}

%% file: introduction.tex

Over the last few years, researchers have proposed many machine learning and deep learning methods to detect online hate speech, which is content that expresses hate or encourages violence towards a person or group based on race, religion, gender, or other identity characteristics \cite{fortuna2018survey,schmidt2017survey}. While many existing methods have achieved reasonably good performance, they have primarily focused on predicting if a given content is hateful with few explanations. In the content moderation process, explanations for flagging are crucial for both moderators and users to understand why a piece of content was deemed hateful. By providing explanations for the predictions, explainable AI can improve the trustworthiness of the hate speech detection methods, increasing the overall accountability of the moderation process \cite{vaccaro2020end,haimson2021disproportionate}.

Previous research has explored various forms of explanations and constructed datasets such as \textit{HateXplain}~\cite{mathew2021hatexplain}, \textit{implicit hate speech} corpus~\cite{elsherief2021latent}, and \textit{social bias frame}~\cite{sap2020social} to support hate speech explanations. The availability of ground truth hate speech explanations can be used to fine-tune or prompt language models such as GPT-3 for free-text explanation generation~\cite{radford2019language}. For instance, \cite{huang2022chain} adopted the Chain-of-thought~\cite{wei2022chain} framework to prompt GPT-2 to generate explanations for hate speech. However, while previous works have performed automatic evaluations by comparing generated explanations to ground truth ~\cite{elsherief2021latent,sap2020social,huang2022chain}, the human perception and implications of these generated explanations remain unclear. This paper aims to evaluate the effectiveness and limitations of the state-of-the-art language model GPT-3 in generating explanations for hate speech. We propose an analytical framework and conduct a comprehensive survey to assess the subjective and nuanced nature of hate speech explanations and answer the following research questions.

\begin{table*}[!ht]
    \centering
    \small
    \begin{tabular}{lp{9cm}ccccc}
        \toprule
        \multicolumn{7}{l}{\textbf{Evaluated Tweet}: 
        ``\textit{maybe ban the nazis before u wanna worry about me tweeting something for the thousandth time}'' [label: Non-hateful]}\\
        \toprule
        & & \multicolumn{5}{c}{\textbf{Human Evaluator Ratings}} \\
        \cmidrule{3-7}
        \textbf{Explanation Type} & \textbf{GPT3 Generated Explanation} & \textbf{F} & \textbf{I} & \textbf{S} & \textbf{P} & \textbf{H} \\
        \midrule

        No explanation. & - & - & - & - & - & [2,2,4]\\
        \midrule
      

        \textsc{Why} (hateful) & This tweet is hateful because it downplays the severity of Nazi ideology and suggests that those who are concerned about it are overreacting... Nazi ideology is responsible for the deaths of millions of people, and it should be taken seriously.
        & 4.0& 3.0& 2.7& 2.0 & [4,1,4]  \\
        \midrule
        \textsc{Why} (non-hateful) & This tweet is non-hateful because it is not making any negative or derogatory statements about Nazis. It is simply making a suggestion that the government should focus on banning Nazis before worrying about what someone is tweeting. & 4.0& 3.3& 3.3& 4.0 & [2,1,1] \\
        \midrule

        \textsc{CoT} (hateful) & 1. The tweet starts with ``maybe ban the nazis." This is clear that the person who wrote this tweet believes that Nazis should be banned. 


        2. This implies that the person who is worried about this tweet is more concerned with preventing the writer from tweeting than they are with banning Nazis.
        
        This tweet is hateful because it downplays the seriousness of Nazis. It also implies that the person who is worried about this tweet is more concerned with censorship than they are with stopping Nazis.
        & 3.0& 1.7& 2.7& 3.0 & [3,4,1] \\
        \midrule
        \textsc{CoT} (non-hateful) & The tweet says ``maybe ban the nazis before u wanna worry about me tweeting ..." This is a suggestion that the person thinks would be more productive than worrying about what the person tweets. 
        
        The person is not making any hateful statements about Nazis. He simply suggests that Nazis should be banned. This is not a hateful statement.
        & 4.0& 4.3& 4.0& 3.7 & [1,1,1] \\
        \midrule
        \textsc{Context} & The tweet is in reference to the recent controversy surrounding Twitter's decision to ban certain users for their hateful and abusive speech. The user is suggesting that Twitter should ban Nazis before they worry about people who are tweeting the same thing over and over again.
        & 4.3& 4.0& 3.7& 3.7 & [3,4,3] \\
        \bottomrule
    \end{tabular}
    \caption{Sample of GPT3 generated explanations and evaluators' 5-point Likert scale rating for the evaluated tweet. The scores are broken down into five categories: \textbf{F} for fluency, \textbf{I} for informativeness, \textbf{S} for soundness, \textbf{P} for persuasiveness, and \textbf{H} for hatefulness. Three annotators assigned each category's ratings. Scores from three annotators were averaged except for hatefulness ratings.
    } 
    \label{tab:motivate_example}
\end{table*}



\begin{enumerate}
    \item \textbf{RQ1}. To what extent do the explanations generated by GPT-3 for hateful content demonstrate qualities of being fluent, informative, persuasive, and logically sound?
    \item \textbf{RQ2}. How persuasive are the GPT-3 explanations, and does the effectiveness of persuasion vary with different prompting strategies?
    \item \textbf{RQ3}. Does the use of GPT-3 explanations lead to incorrect decisions in hateful content moderation?
\end{enumerate}


The study utilizes the tweets annotated in the HateXplain dataset~\cite{mathew2021hatexplain}. The study aims to prompt GPT-3 to explain hateful and non-hateful tweets and assess the generated explanations' quality through human evaluations. The evaluation metrics include fluency, informativeness, soundness, and persuasiveness. We conducted a survey to assess the quality of explanations generated by GPT-3 and the hatefulness of tweets after viewing the explanations. The study addresses three key research questions, and we summarize the key findings as follows:


\begin{itemize}
    \item Human evaluators have assessed that the explanations generated by GPT-3 are fluent, informative, persuasive, and logically sound.
    \item Different prompting strategies illicit varying persuasive effects. When prompted to explain why a given tweet is hateful, GPT-3 generated a more persuasive explanation than simply asking it to provide contextual information on the tweet. The length of the generated explanation also affects its persuasiveness.
    \item The potential for GPT-3 generated explanations to mislead human moderators when evaluating hateful content is a matter of concern. Our study has revealed that the explanations produced by GPT-3 can cause human evaluators to misclassify roughly 20\% of tweets. Such misclassifications can wrongly label non-hateful tweets as hateful or vice versa. This is in contrast to a baseline scenario where human evaluators made assessments without any explanation.
    \item We observed that presenting both hateful and non-hateful explanations generated by GPT-3 could mitigate the risk of misleading content moderators.

    
    
\end{itemize}

%% file: related.tex
\paragraph{Content moderation and hate speech.} In recent years, major social media platforms such as Facebook, YouTube, and Twitter have integrated AI into their operations for content moderation. The rise of user-generated content has made it increasingly necessary for these platforms to monitor and remove harmful or illegal content~\cite{djuric2015hs,badjatiya2017hs,watanabe2018hs,awal2021angrybert,awal2023model,meng2022predicting,lin2021early}. As a result, governments have imposed strict regulations on social media platforms, requiring them to remove hateful content quickly. In response, many platforms have implemented automated systems, such as algorithms and AI, to proactively detect and remove such content at scale~\cite{lampe2004slash}. However, these systems have been criticized for labour concerns, lack of transparency, perpetuating biases, and potential harm to marginalized communities~\cite{gillespie2018custodians,haimson2021disproportionate,steiger2021psychological,suzor2019we,cao2020hategan}. To address these concerns, there is a growing emphasis on developing trustworthy and explainable AI systems for hate speech detection and moderation. 

\paragraph{Explainable hate speech detection.} Hate speech detection aims to identify and prevent online hate. Although several studies have been conducted on this topic~\cite{davidson2017automated,cao2020deephate}, hate speech models like the Google Jigsaw API are vulnerable to racial biases and counterfactual and adversarial attacks~\cite{sap2019risk,park2018reducing}. To address these issues, explainable hate speech detection approaches have received attention, wherein model predictions are described through natural language explanations~\cite{camburu2018snli,mathew2021hatexplain}. Recent studies, such as \cite{sap2020social} and \cite{elsherief2021latent}, have examined stereotypes in social media and the implicit nature of hate speech, respectively, providing insight into the underlying causes of hate speech and aiding in the development of model explanations. Furthermore, the generation of counter-narratives to combat hate speech has been explored~\cite{tekiroglu2020cn,ashida2022towards}.

Human-generated explanations are limited and challenging to scale for real-world scenarios. To overcome this limitation, pre-trained language models are being utilized to automate the explanation generation process~\cite{sap2020social,huang2022chain}. Furthermore, generative explanations can help address user appeals against platform flagging~\cite{vaccaro2020end}. However, using GPT-3 based API for content moderation by companies like Cohere and OpenAI raises concerns about bias and spreading misinformation~\cite{markov2022holistic}. Additionally, the quality of generated explanations requires further evaluation, as current studies primarily rely on automated quantitative analysis, such as the BLEU score. Therefore, our research aims to assess the credibility of generated explanations using a qualitative assessment conducted by human evaluators.


\paragraph{Prompting.} Prompting is a technique that enables LLMs, such as GPT-3, to adapt to specific tasks by incorporating task-related instructions or questions into input text~\cite{brown2020language,sanh2021multitask,wei2022chain,kojima2022chain}. The format and order of the prompt and the demonstration examples, can affect the performance of LLMs~\cite{zhang2022chain,zhao2021chain}. Complex reasoning tasks can be improved by inducing a sequence of intermediate steps with few human-crafted demonstration examples~\cite{wei2022chain}. Adding the instruction "Let's think step by step" before each response could have the same effect as~\cite{wei2022chain} in zero-shot learning~\cite{kojima2022chain}. Prompting has shown promising results in various natural language processing (NLP) tasks, including teaching machines to generate explanations with demonstrations, which can be even more suitable than human-written explanations~\cite{wiegreffe2022reframing,petroni2019language}. 

Previous research has investigated using GPT-3 for addressing online hatred and abuse, including hate speech detection~\cite{chiu2021lms} and toxicity detection~\cite{wang2022lms}. GPT-3 has the potential to generate explanations for its predictions, such as why it classified certain texts as hate speech. However, the generated content from GPT-3 may not always be accurate or factual, presenting potential risks in applications such as addressing online hatred and abuse. Generative models may not always critically evaluate their predictions, meaning they can make predictions but not provide reasoning behind them~\cite{saunders2022self}. GPT-3's ability to generate human-like text can be misused to create convincing disinformation, such as fake news and propaganda. Its writing has been shown to significantly impact readers' perspectives on international affairs~\cite{buchanan2021truth}. It is crucial to weigh the potential benefits of using GPT-3 against the risks associated with its ability to generate convincing disinformation and influence people's perspectives, as well as its limitations in providing reasoning for its predictions.

%% file: method.tex
\begin{figure}[t]
  \includegraphics[width=0.45\textwidth]{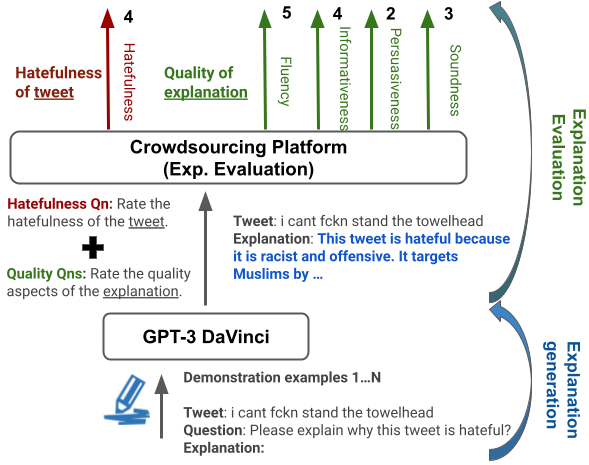}
    \caption{Overall framework of our study depicting the procedure of generating explanations utilizing GPT-3 and conducting a human evaluation. To prompt GPT-3, we selected demonstration examples from a candidate pool, producing one explanation for each tweet, based on each prompting strategy. Subsequently, human evaluators, via the mTurk and Clickworker platforms, assessed tweets' hatefulness and the explanations' quality. The outcomes of the human evaluation process were utilized to appraise the effectiveness of the explanations.}
      \label{fig:experiment_flow}
\end{figure}

This section outlines our methodology for determining the quality of explanations generated for hate speech using a representative dataset. The simplified process is as shown in Figure~\ref{fig:experiment_flow}. The overall process includes sampling examples of hate speech, utilizing various prompting strategies to generate explanations, and designing a survey to evaluate the quality of the generated explanations. The survey is administered to gain deeper insights into the nature of hate speech and how to combat it effectively. The following sections detail our methods and techniques used at each stage of the process.

\subsection{Dataset}

The \textbf{HateXplain} dataset~\cite{mathew2021hatexplain} comprises a set of hate speech instances acquired from Twitter and Gab. Each record in the dataset was annotated by three persons who assigned one of the following three labels: "offensive," "normal," or "hateful." The final label for each record was determined via majority voting. Our experiment only considered Twitter tweets labelled "hateful" or "normal." To ensure a comprehensive analysis of hate speech, we incorporated all annotations associated with each tweet. Consequently, the tweets were categorized into four groups: [nnn, nnh, nhh, hhh], where "h" represents "hateful" and "n" represents "normal." A total of 25 tweets were randomly selected from each group, resulting in a 100-tweet evaluation set.

\subsection{Prompting GPT-3 for Explanation Generation}
We fed the GPT-3 model with the evaluated tweet, followed by three prompting strategies to generate the explanations:

\begin{itemize}
    \item \textbf{Why is ...? (\textsc{WHY})} We prompt the model with ``\textit{Please explain why this tweet is (hateful/non-hateful)}''. The goal is to prompt GPT-3 to generate a hateful or non-hateful explanation for the tweet.
    \item \textbf{Chain-of-Thought (\textsc{CoT})}: We prompt the model with the same question asked in the \textsc{WHY} strategy and prefix the answer ``\textit{Let's think step-by-step.}''. The goal is to prompt GPT-3 to generate a more detailed hateful or non-hateful explanation with reasoning steps.
    \item \textbf{What is the context ...? (\textsc{CONTEXT})}: We prompt the model with ``\textit{What is the context of the tweet?}". The goal is to encourage the GPT-3 model to generate an informative response that captures the relevant contextual details associated with the tweet without explicitly telling the model if the tweet is hateful or non-hateful.
\end{itemize}

We prompted GPT-3 with four demonstration examples for each strategy before asking about the specific tweet.
These examples consist of tweets extracted from the HateXplain dataset, accompanied by explanations generated through zero-shot prompting using GPT-3 and carefully chosen for their fluency, soundness, informativeness, and persuasiveness. The Table~\ref{fig:prompt_design} presents the overall prompt format.  

Using the three prompting techniques, eight distinct combinations of tweets and explanations were generated for survey evaluation. These combinations involved pairing a tweet with either no explanation, a hateful explanation, a non-hateful explanation, a contextual explanation, or both hateful and non-hateful explanations side-by-side.


\definecolor{question_color}{HTML}{1B9E77}
\definecolor{label_color}{HTML}{D95F02}
\definecolor{explanation_provided_color}{HTML}{7570B3}
\definecolor{explanation_generated_color}{HTML}{FF55A3}
\definecolor{tweet_color}{HTML}{FFC300}

\begin{table}[t!]
\centering
\small
\begin{tabular}{p{0.90\linewidth}}
\toprule
\# Demonstration samples $1 ... N$\\
\textbf{Tweet}: \textcolor{tweet_color}{$<$user$>$ $<$user$>$ $<$user$>$ $<$user$>$ $<$user$>$ what else we can expect from jihadi terrorist country}\\
\textbf{Question}: \textcolor{question_color}{Please explain why this tweet is hateful?}\\
\textbf{Answer}: \textcolor{explanation_provided_color}{[Let's think step-by-step.] The tweet is hateful because it mocks Pakistan's status as a haven for terrorists.}\\
\#\#\#\\

\# Actual tweet in prompt\\
\textbf{Tweet}: \textcolor{tweet_color}{$<$ser$>$ so did the nazis,}\\
\textbf{Question}: \textcolor{question_color}{Please explain why this tweet is hateful?}\\
\textbf{Answer}: \textcolor{explanation_provided_color}{[Let’s think step-by-step.]} \textcolor{explanation_generated_color}{This tweet is hateful because it trivializes the Holocaust and the suffering of the Jewish people. It is also offensive and insensitive.}\\
\bottomrule
\end{tabular}
\caption{An illustration of a prompt utilizing $N$ training examples is presented, consisting of the tweet (in \textcolor{tweet_color}{yellow}), the question (in \textcolor{question_color}{green}), and the explanations provided (in \textcolor{explanation_provided_color}{purple}) that are utilized as input to GPT-3.}
\label{fig:prompt_design}
\end{table}


\begin{table*}[!t]
\small
    \centering
    \begin{tabular}{lcccccccc}
       \toprule
        & & \multicolumn{3}{c}{\textbf{\textsc{Why}}} & \multicolumn{3}{c}{\textbf{\textsc{CoT}}} & \textbf{\textsc{CONTEXT}} \\
        \cmidrule(l{2pt}r{2pt}){3-5}\cmidrule(l{2pt}r{2pt}){6-8}
       & \textbf{Tweets label} & \emph{hateful} & \emph{non-hateful} &\emph{both} & \emph{hateful} & \emph{non-hateful} &\emph{both}\\
       \midrule
       \multirow{2}{*}{\centering \textbf{Fluency}} & not-hate & 3.91	& 3.33 & 3.92 & 3.63 & 3.65 &3.6 &3.07 \\
       \arrayrulecolor{black!20}\cmidrule{2-9}
        & hate & 3.85	&3.04 &3.9	&3.65	&3.42&3.49&2.39 \\
        \arrayrulecolor{black}\hline

        \multirow{2}{*}{\centering \textbf{Informativeness}} & not-hate & 3.65	&3.21 &3.39	& 3.55	& 3.65 &3.51&2.93 \\
       \arrayrulecolor{black!20}\cmidrule{2-9}
        & hate & 3.75	&2.59 &3.38	&3.51	&3.39 &3.35&1.95\\
        \arrayrulecolor{black}\hline

        \multirow{2}{*}{\centering \textbf{Persuasiveness}} & not-hate & 3.37	& 3.03 &3.2	& 3.27	& 3.21 &3.21&2.71 \\
       \arrayrulecolor{black!20}\cmidrule{2-9}
        & hate & 3.6	&2.33	&3.19& 3.38	&2.79&3.02&1.86\\
        \arrayrulecolor{black}\hline

        \multirow{2}{*}{\centering \textbf{Soundness}} & not-hate & 3.49	& 3.14	&3.2&3.47	& 3.41&3.41&3.09	\\
        \arrayrulecolor{black!20}\cmidrule{2-9}
        & hate & 3.7	&2.35	&3.35& 3.65	& 2.88 &3.26&2.16\\
       \arrayrulecolor{black}\bottomrule
    \end{tabular}
    \caption{Quality assessment of the \textsc{Why}, \textsc{CoT}, and \textsc{Context} prompting strategies for generating GPT-3 explanations for hateful and non-hateful tweets. It is worth noting that for every tweet, irrespective of its label, we utilized the \textsc{Why} and \textsc{CoT} strategies to prompt for three types of explanations: \textit{hateful}, \textit{non-hateful}, and \textit{both}, where both hateful and non-hateful explanations were generated side-by-side.}
    \label{tab:quality_score}
\end{table*}

\subsection{Evaluation Metrics}
An extensive human evaluation was conducted to evaluate the hatefulness of the tweets and the quality of the explanations generated by GPT-3 based on four criteria: \textit{fluency}, \textit{informativeness}, \textit{persuasiveness}, and \textit{soundness}. A 5-point Likert scale was employed, where 1 represented the poorest quality and 5 the best. The definition of hatefulness and the quality criteria are as follows.
\begin{itemize}[leftmargin=*]
    \item \textbf{Hatefulness.} A \textit{hateful} tweet is defined as: ``Any speech that attacks a person or group on the basis of attributes such as race, religion, ethnicity, nationality, gender, disability, sexual orientation, or other identity factors." The rating scale spanned from 1 (non-hateful) to 5 (hateful), allowing annotators to express varying degrees of hate speech \cite{poletto2019annotating}. This approach offers a more detailed and flexible annotation process.
    \item \textbf{Fluency} evaluates whether the explanation follows proper grammar and structural rules, with a rating scale ranging from 1 (poor) to 5 (excellent).
    \item \textbf{Informativeness} assesses whether the explanation provides new information, such as explaining the background and additional context, with a rating scale ranging from 1 (not informative) to 5 (very informative).
    \item \textbf{Persuasiveness} evaluates whether the explanation seems convincing, with a rating scale ranging from 1 (not persuasive) to 5 (very persuasive).
    \item \textbf{Soundness} describes whether the explanation seems valid and logical, with a rating scale ranging from 1 (not sound) to 5 (very sound).
\end{itemize}

\subsection{Human Evaluation Setting}
Human evaluations were carried out on both Amazon Mechanical Turk and Clickworker platforms. As the HateXplain dataset primarily comprises Twitter content generated in a American context, we recruited human evaluators residing in the United States. In total, we recruited 2,400 participants for the human evaluation. They were asked to evaluate the level of hatefulness in tweets and the quality of explanations generated by the GPT-3 model.

A three-round survey was executed to evaluate the explanations generated by GPT-3. Each round involved distinct participants and a variable number of questions pertaining to the same set of 100 sampled tweets. To ensure the validity of the survey and the respondent attentiveness, a basic math question is included in each round, e.g. ``What is 8+3?''.

\paragraph{Round 1.} In the survey's first round, participants assessed tweet hatefulness without explanations. This controlled baseline aimed to determine if providing explanations impacts hatefulness perception in those tweets.

\paragraph{Round 2.} In the second round of the survey, tweet-explanation pairs were presented, with each tweet paired with either a hateful or a non-hateful explanation. The explanations were generated using the prompting strategies \textsc{Why} and \textsc{CoT}. Following this, participants were asked to evaluate the level of hatefulness in the tweets after reading the provided explanation and respond to four quality questions assessing the fluency, informativeness, persuasiveness, and soundness of the explanation. The primary objective of this round was to examine whether GPT-3 can generate high-quality and persuasive explanations (for both hate and non-hate labels) capable of influencing individuals' perceptions.

\paragraph{Round 3.} The third round of the survey adopts two distinct formats. The first format combines hateful and non-hateful explanations generated by the \textsc{Why} and \textsc{CoT} prompting strategies. Participants were asked to respond to four quality questions of both explanations presented side-by-side. In the second format, the \textsc{Context} prompting strategy was utilized to elicit background information about the given tweet. Based on the contextual information, participants had to answer four quality questions assessing the explanations' quality. The primary objective of this format is to evaluate whether language models can generate high-quality and objective explanations (contextual) that reveal the tweet's implicit context without impacting readers' perceptions.

Participants who did not receive any explanations (i.e., Round 1) were given a survey consisting of two questions and were allotted 15 minutes to complete it. Conversely, participants who were presented with one explanation (i.e., Round 2) were given a survey consisting of five questions and given 30 minutes to complete it, while those who received two explanations (i.e., Round 3) were provided with a survey consisting of 10 questions and given 45 minutes to complete it. Three survey responses were collected for each data point, and only responses from evaluators who correctly answered the basic arithmetic questions were recorded in the analysis. 


%% file: experiments.tex
This section presents the human evaluation results and discusses our key findings in respect to the three research questions on hate speech explanation generated by GPT-3.


\subsection{RQ1. Quality of Generated Explanations}
\label{sec:rq1}

\begin{table*}[t]
    \centering
    \adjustbox{max width=\linewidth}{
    \begin{tabular}{cp{1.2cm}|ccccccc}
       \toprule
        & & \multicolumn{3}{c}{\textbf{\textsc{Why}}} & \multicolumn{3}{c}{\textbf{\textsc{CoT}}}  & \textbf{\textsc{Context}} \\
        \cmidrule(l{2pt}r{2pt}){3-5}\cmidrule(l{2pt}r{2pt}){6-8}
       \centering \textbf{Tweets label} & \textbf{No Exp.} & \emph{hateful} & \emph{non-hateful} & \emph{both} &\emph{hateful} & \emph{non-hateful}   & \emph{both}  & \\
       \hline
       
        non-hate & 3.16 & \textcolor{red}{$3.64(+0.48)^{**}$} & \textcolor{blue}{$2.73(-0.43)^{**}$} & \textcolor{blue}{3.14(-0.02)} &	\textcolor{red}{3.25(+0.09)} &	\textcolor{blue}{$2.69(-0.47)^{**}$} &	\textcolor{blue}{3.13(-0.03)}&	\textcolor{blue}{$2.82(-0.24)^{*}$} \\ 
       \arrayrulecolor{black!20}\cmidrule{1-9}
        
       hate & 3.96 & \textcolor{blue}{3.94(-0.02)} & \textcolor{blue}{$3.57(-0.39)^{**}$}  & \textcolor{red}{3.97(+0.01)} &	\textcolor{red}{$4.39(+0.43)^{**}$} &	\textcolor{blue}{3.81(-0.15)}  & \textcolor{red}{4.11(+0.15)} &	 \textcolor{red}{4.05(+0.11)} \\ 
       \arrayrulecolor{black}\bottomrule
    \end{tabular}
    }
    \caption{Average \textit{hatefulness} scores of the tweets evaluated by human evaluators after viewing  explanations generated by GPT-3 using different prompting strategies. The values in () represent the differences between the tweets' average \textit{hatefulness} scores without viewing any explanations (i.e., No Exp.) and after viewing explanations. \textcolor{red}{Red} indicates an increase in \textit{hatefulness}, while \textcolor{blue}{blue} indicates a decrease in hatefulness. To determine the significance of the differences, we computed p-value, denoted by ** when $p \leqslant 0.01$ and * when $p \leqslant 0.05$. }
    \label{tab:hatefulness}
\end{table*}

Table~\ref{tab:quality_score} summarizes the quality assessment of the GPT-3 generated explanations using various prompting strategies. Specifically, we report the average \textit{Fluency}, \textit{Informativeness}, \textit{Persuasiveness}, and \textit{Soundness} scores provided by the human evaluators. These were rated on a scale of 1 to 5, with 5 being the best score. 

We observe that within the same prompting strategy, the scores across four quality metrics for hateful explanations were higher than that for non-hateful explanations. This is because the HateXplain dataset used in our study focuses on hate speech and the tweets collected were more offensive in nature. Some tweets contain slurs such as ``\textit{niggas}",  ``\textit{faggots}" or ``\textit{fuck}", but are marked as non-hateful because of its context. Consequently, GPT-3 faces a greater challenge in generating reasonable non-hateful explanations for such tweets, ultimately impacting human evaluators' assessments. We compared two strategies, \textsc{Why} and \textsc{CoT}, for generating non-hateful explanations and found that the latter significantly improved the quality score (with an average improvement of approximately +0.42 across all quality assessment metrics). The superior performance of the \textsc{CoT} approach in generating non-hateful explanations suggests that breaking down the tweet analysis into steps to reach a conclusion is beneficial in generating non-hateful explanations, particularly in challenging contexts such as those presented by the HateXplain dataset. Conversely, we observe that the hateful explanations generated using \textsc{Why} strategy scored better than those generated using \textsc{CoT}, suggesting that the human evaluators prefer shorter and more direct hateful explanations over long and elaborate ones when evaluating generally offensive tweets.

Our findings indicate that the quality scores for the side-by-side hateful and non-hateful explanations (\textit{both}) generated using \textsc{Why} and \textsc{CoT} were moderately effective. In contrast, \textsc{Context} was the least effective prompting strategy for generating high-quality explanations. Closer examination of the \textsc{Context} explanations revealed that many of them simply summarized the tweet content without providing significantly new information. In Section~\ref{sec:rq3}, we will delve deeper into the effects of applying the \textsc{Why}-\textit{Both}, \textsc{Cot}-\textit{Both}, and \textsc{Context} strategies to mitigate risks associated with using LLMs for moderating hateful content.

Overall, our quality assessment analysis revealed that the explanations generated by GPT-3 scored well in terms of \textit{fluency} and \textit{informativeness}, but slightly worse in terms of \textit{soundness} and \textit{persuasiveness}. This finding is consistent with previous research \cite{wiegreffe2022reframing,dou2021gpt3}, which suggests that generating a sound and persuasive explanation necessitates a deeper understanding of the tweets, making it a more challenging task than achieving fluency and informativeness. Furthermore, we observed that the GPT-3 model performed inadequately in generating explanations with a prompting label that contradicts the true label in the dataset. Consequently, hateful explanations received higher quality scores for hateful tweets than non-hateful tweets, while non-hateful explanations received higher scores for non-hateful tweets than hateful tweets.

\subsection{RQ2. Persuasiveness of Generated Explanations}
\label{sec:rq2}
The earlier section has established that GPT-3 is able to generate high quality explanations for hateful tweets. However, little is known about the the influence of these generated explanations on human perception. This section will fill this gap by analyzing the persuasiveness of the generated explanations and their effects on the the tweets' \textit{hatefulness} scores rated by the human evaluators. 

Table~\ref{tab:hatefulness} presents a summary of the average \textit{hatefulness} scores of tweets rated by human evaluators when presented with the accompanying explanations generated by GPT-3 using various prompting strategies. We observe that the \textsc{why}-\textit{non-hateful} explanations have significantly decreased the \textit{hatefulness} scores of both hateful and non-hateful tweets. Conversely, we observe that the \textsc{why}-\textit{hateful} explanations have significantly increased the \textit{hatefulness} scores of non-hateful tweets, but no such effects are observed on the hateful tweets.  One possible explanation for this finding is that the human evaluators had already rated the hateful tweets as hateful, and the generated explanation did not alter their assessment of these tweets.

Similar observations were made on the explanations generated using the \textsc{CoT} strategy. For instance, the \textsc{CoT}-\textit{non-hateful} explanations have significantly decreased the \textit{hatefulness} scores of non-hateful tweets, and the \textsc{CoT}-\textit{hateful} explanations have significantly increased the \textit{hatefulness} scores of hateful tweets. Interestingly, we observe that the \textsc{CoT}-\textit{non-hateful} explanations did not significantly change the \textit{hatefulness} scores of hateful tweets. This finding may be attributed to the elaborated explanations generated by the \textsc{CoT} strategy, which may have required the GPT-3 model to fabricate convoluted explanations in an attempt to convince human evaluators that a hateful tweet is actually non-hateful. However, as discussed in the previous section, such explanations may be deemed as unsound, thereby failing to alter the human evaluators' ratings of the tweet. Similar results were observed for the \textsc{CoT}-\textit{hateful} explanations on non-hateful tweets.

When presented with both hateful and non-hateful explanation side-by-side using the \textsc{why}-\textit{both} and \textsc{CoT}-\textit{both} strategies, we observe that \textit{hatefulness} scores have little or no apparent change compared to not providing any explanations. This suggests that when presented with a balanced explanation, i.e., both hateful and non-hateful, the human evaluators are less likely to be persuaded by either explanation and alter their opinions on a given tweet. Similarly, when presented with the \textsc{context} explanations for hateful tweets, the change in \textit{hatefulness} scores is insignificant. Interestingly, we noted that providing the \textsc{context} explanations reduced the hatefulness scores for non-hateful tweet significantly.

\subsection{RQ3. Risks of GPT-3 Generated Explanations for Hateful Content Moderation}
\label{sec:rq3}

\begin{figure}[t!]
  \includegraphics[width=0.48\textwidth]{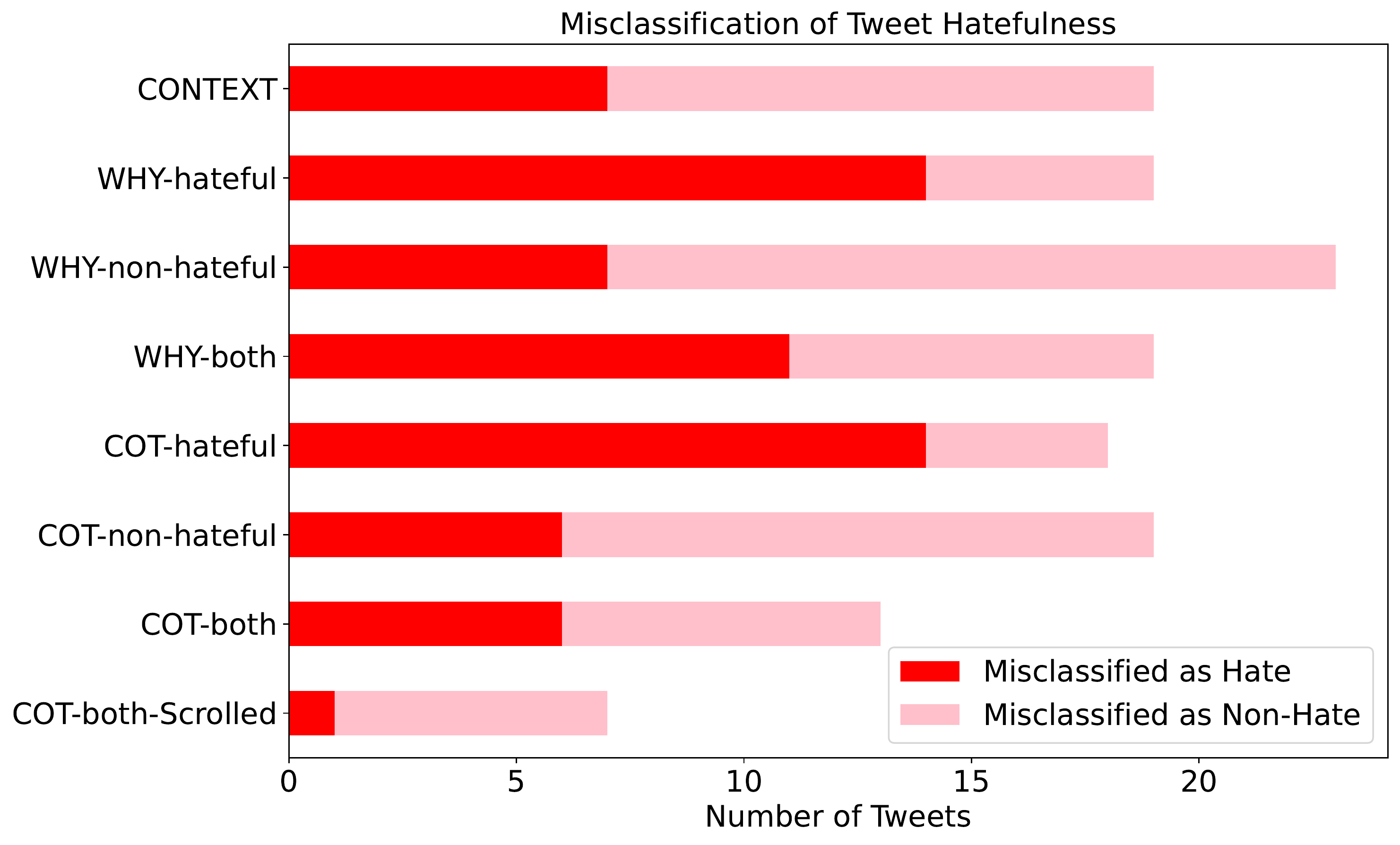}
    \caption{Distribution of tweets misclassified by human evaluators after reading prompting strategies' explanations.}
      \label{fig:misclassification}
\end{figure}









In this analysis, we aim to examine the impact of explanations generated by GPT-3 on human decision-making in hateful content moderation. To achieve this, we first classify each annotation as either \textit{non-hateful} or \textit{hateful}, based on their rated hatefulness scores of 1-2 and 4-5, respectively. Each tweet has three annotations from three different human evaluators, and the majority vote is used as the annotated label for the tweets. We identified misclassifications by comparing the labels assigned by human evaluators with and without seeing explanations. Specifically, we studied cases where human evaluators labeled tweets different from round 1, where there are no explanations, after reading GPT-3 generated explanations using various prompting strategies. Figure~\ref{fig:misclassification} displays the distributions of tweets that were misclassified by human evaluators after reading explanations generated by different prompting strategies.

\paragraph{Misleading Human Evaluators.} Figure~\ref{fig:misclassification} shows the distribution of tweets misclassified by the human evaluators after reading the prompting strategies' explanation. From Figure~\ref{fig:misclassification}, we observe that exposing the evaluators to \textsc{WHY}-\textit{hateful} explanations increases the misclassification of non-hateful tweets, as they are persuaded to label them as hateful. Similarly, exposing the evaluators to \textsc{WHY}-\textit{non-hateful} explanations also leads to an increase in misclassifying hateful tweets as non-hateful. The \textsc{COT}-\textit{hateful} and \textsc{COT}-\textit{non-hateful} explanations also misled the evaluators, resulting in misclassification of both non-hateful and hateful tweets. Surprisingly, even \textsc{Context} explanations have been found to mislead evaluators, resulting in misclassifications of both hateful and non-hateful tweets. Notably, a significant number of hateful tweets were mistakenly rated as non-hateful when presented with \textsc{Context} explanations.

Our hypothesis was that presenting both hateful and non-hateful explanations together would provide human evaluators with balanced information, aiding them in making better decisions regarding moderating hateful content. However, our observations show that even with \textsc{WHY}-\textit{both} explanations, there is still a significant number of misclassifications. Conversely, our findings indicate that the \textsc{COT}-\textit{both} approach is a more promising method, resulting in fewer misclassifications for both hateful and non-hateful tweets.

\paragraph{Effects of Explanation Length.} Despite the ability of \textsc{COT}-\textit{both} to provide detailed and balanced explanations to human evaluators, we found that the generated explanations are significantly longer than those generated by other methods. For example, \textsc{COT}-\textit{both} explanations have an average length of 173 words, compared to \textsc{Context} and \textsc{Why}-\textit{both} explanations with an average length of 28 and 75 words, respectively. Additionally, our research revealed that when human evaluators used a mobile device to respond to the survey, some \textsc{COT}-\textit{both} explanations extended beyond a single page, requiring the evaluator to scroll to view entire content. To address this issue, we introduced a ``scrolled" variable in our survey to verify whether evaluators had scrolled to read the full \textsc{COT}-\textit{both} explanations when rating the hatefulness of the tweets. 
We next analyzed the scrolling behavior of the 300 evaluators who assessed tweets paired with \textsc{COT}-\textit{both} explanations. We discovered that 69 out of the 300 evaluators did not scroll to read the complete \textsc{COT}-\textit{both} explanations while rating the hatefulness of the tweets. This raises concerns about the reliability of these evaluations, and we therefore excluded these 69 ratings from our analysis. The resulting misclassification of tweets is reported in Figure~\ref{fig:misclassification} as \textsc{COT}-\textit{both Scrolled}. The figure shows that the number of misclassifications decreased even further after removing the evaluations of human raters who did not scroll to read the full \textsc{COT}-\textit{both} explanations. Misclassifications may occur when individuals focus solely on the first hateful explanations. This selective exposure raises the risk of mistakenly classifying non-hate tweets as hate and the drawback of lengthy explanations is also evident. Hence, future research will focus on exploring techniques that generate shorter and more concise explanations for effective moderation of hateful content.

%% file: discussion.tex

To the best of our knowledge, this study represents the first evaluation of the quality and ethical implications of PLM-generated explanations for moderating hateful content on a large scale. Our findings demonstrate that GPT-3 can produce fluent, informative, persuasive, and logically sound explanations when properly prompted. These results suggest that GPT-3 may be a valuable tool for combating online hate speech and content moderation. However, we discovered that the persuasiveness of the explanations varied depending on the prompting strategy and the length of the explanations. In particular, we found that biased explanations, regardless of whether they were hateful or non-hateful, had a strong persuasive effect on human annotators. Therefore, the use of GPT-3 to produce explanations for content moderation should be approached with caution. Moreover, our observations revealed that the explanations generated by GPT-3 had the potential to lead to incorrect labeling of tweets. This is a critical issue that must be addressed to ensure the effectiveness and ethicality of using GPT-3 to aid content moderation. We recommend presenting both hateful and non-hateful explanations side-by-side, which can significantly reduce the risk of misleading content moderators.

\paragraph{Limitations.} The study analyzed the Hatexplain dataset, which specifically addresses hate speech on Twitter and Gab, which may limit the generalizability of the study to other platforms. Moreover, the Hatexplain dataset, like other hate speech detection datasets, is susceptible to subjective hate speech annotation, leading to varying definitions among annotators and potentially inaccurate annotations. Additionally, the study used non-expert human annotators to evaluate the GPT-3 model's explanations, with precautions taken for accuracy, but some anomalies may still exist. Future research could involve experienced human annotators to obtain higher-quality annotations. The study solely concentrated on the GPT-3 model, necessitating further investigation into whether similar results can be achieved with other language models.

%% file: conclusion.tex
This study introduced an analytical framework for evaluating hate speech explanations and conducted a comprehensive survey on their effectiveness. The research identified the potential of GPT-3 in generating high-quality explanations for hate speech while also highlighting the limitations and risks of using PLM-generated explanations for content moderation. For future work, we aim to develop improved evaluation metrics and prompting strategies to enhance the quality and reduce bias in explanations. These efforts would help increase the fairness and effectiveness of content moderation and combat online hate speech.
